# Sentiment Analysis on Social Media Content

Antony Samuels

University of Southern California

aantonysamuels@gmail.com

John Mcgonical

Caltech

to.john.mcgonical@gmail.com

***Abstract*— Nowadays, people from all around the world use social media sites to share information. Twitter for example is a platform in which users send, read posts known as 'tweets' and interact with different communities. Users share their daily lives, post their opinions on everything such as brands and places. Companies can benefit from this massive platform by collecting data related to opinions on them. The aim of this paper is to present a model that can perform sentiment analysis of real data collected from Twitter. Data in Twitter is highly unstructured which makes it difficult to analyze. However, our proposed model is different from prior work in this field because it combined the use of supervised and unsupervised machine learning algorithms. The process of performing sentiment analysis as follows: Tweet extracted directly from Twitter API, then cleaning and discovery of data performed. After that the data were fed into several models for the purpose of training. Each tweet extracted classified based on its sentiment whether it is a positive, negative or neutral. Data were collected on two subjects McDonalds and KFC to show which restaurant has more popularity. Different machine learning algorithms were used. The result from these models were tested using various testing metrics like cross validation and f-score. Moreover, our model demonstrates strong performance on mining texts extracted directly from Twitter.**

## I. Introduction

The online social media such as Twitter, Facebook, and Instagram allow users to communicate with the whole world. Write their own opinions about products or share their moments, even influence politics and companies. Twitter for example, almost every huge company have an account on Twitter to know about their customers feedback about their services or products. Sentiment analysis, known as opinion mining, for classifying specific words into positive or negative. [1-4].

In this paper, we used sentiment analysis to classify specific English tweets about two restaurants, KFC and McDonald's. our research was determining which one better than other, in specific we examined weather specific tweets is positive, negative, neutral.

## II. Related Work

In [5] gives a focus on analyzing tweets in written English language, belong to different telecommunication companies of KSA, for performing opinion mining on it, they used a supervised machine learning algorithms for classification. Moreover, they used TF-IDF (Term frequency – inverse document frequency) to measure how important word is to a specific tweet. In [6] develops sentiment analysis approach embedded in public Arabic tweets and Facebook comments. They used supervised machine learning algorithms such as Support Vector Machine (SVM) and Naïve Bayes, and they used binary model(BM) and TF-IDF to see the effect of several terms weighting functions on the accuracy of sentiment analysis. In [7] applies sentiment analysis on twitter dataset of 4700 for Saudi dialect sentiment analysis with (k=0.807), they used natural language analysis for Arabic language text. Researchers in [8] present a sentiment analysis for Egyptian dialect using a corpus such as tweets, products review data. They apply natural language processing to understand the Egyptian dialect. Additionally, they used a lexicon-based classification to classify the data.

## III. Methodology

This paper focus on mining tweets written in English. We are interested in seeing who people think is better Mcdonalds or KFC in terms of how good/bad reviews are. Analyzing people's opinions and what they think about a product from their tweets on social media could be a valuable thing for any business. In our project, we extracted tweets from Twitter using R language. R is a programming language used for statistical computing and machine learning algorithms. In order to extract tweets from Twitter, Twitter API were used to create Twitter application and get authorization. In Rstudio which is an environment and graphical user interface for R, we installed necessary packages and libraries. Some of the packages are (TwitteR, rtweet, R0Auth). By using twitteR package you can extract tweets up to 4000 only [9-12].

On the other hand, there is retweet package which is way better than twitteR because it allows you to extract up to 20,000 tweets and in a suitable format (see Table I).

TABLE I. NUMBER OF TWEETS EXTRACTED

| ***McDonald's*** | ***KFC*** |
|---|---|
| 7000 tweets | 7000 tweets |

Before cleaning and preparing the tweets, we want to explore the data and get insight into it. Fig. 1 and Fig. 2 shows the frequency of tweets during the day for both McDonalds and KFC.

From Fig. 1 we found that people tend to tweet about McDonalds at different times during the day, as we can see the lowest number of tweets was during the morning in the interval from 6AM till 12PM. It is different from day to day which is interesting.

From Fig. 2 we see that people tend tweet about KFC at different times during the day, but the lowest number of tweets was during the morning around 10AM.

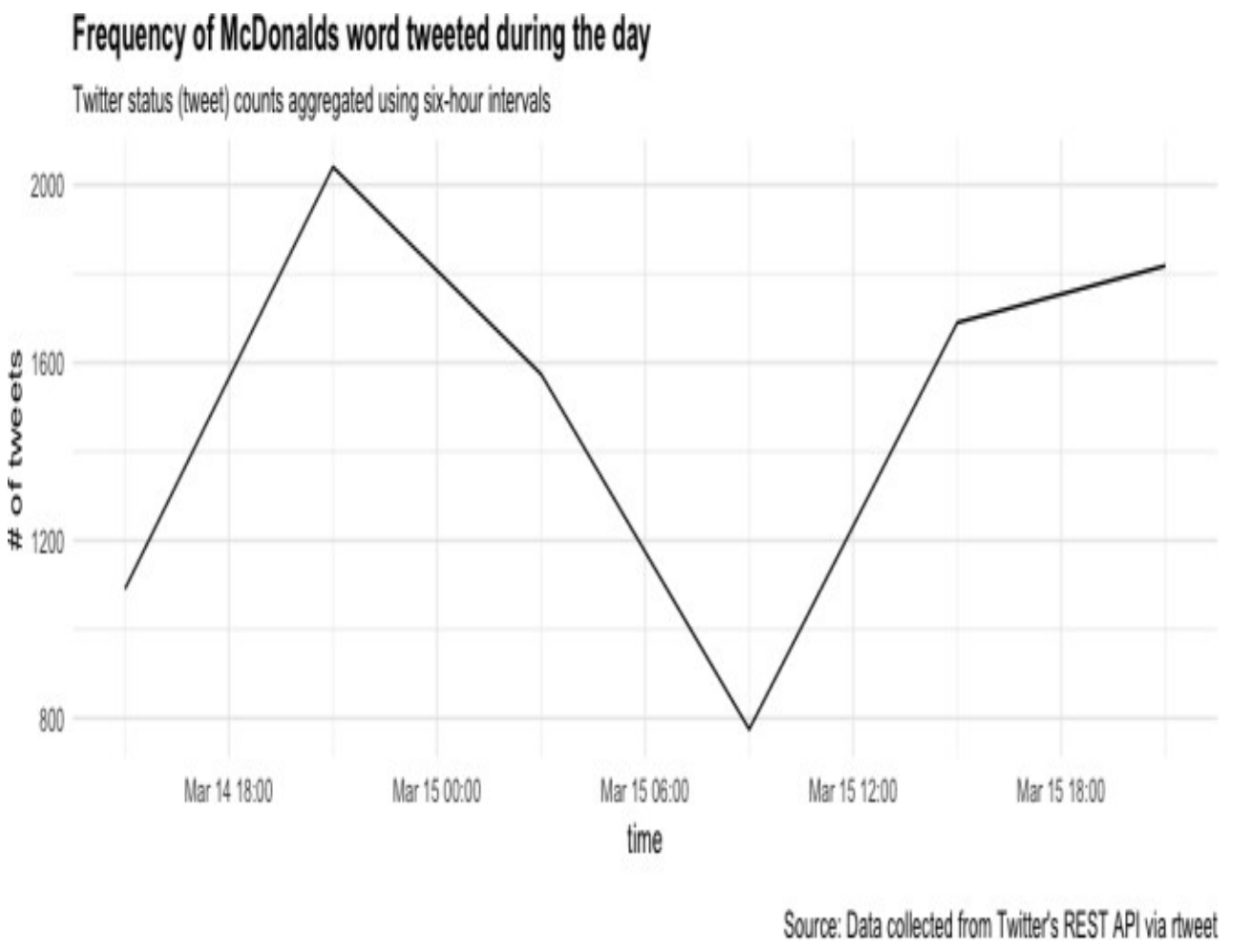

Fig. 1. Frequency of tweets about McDonald's.

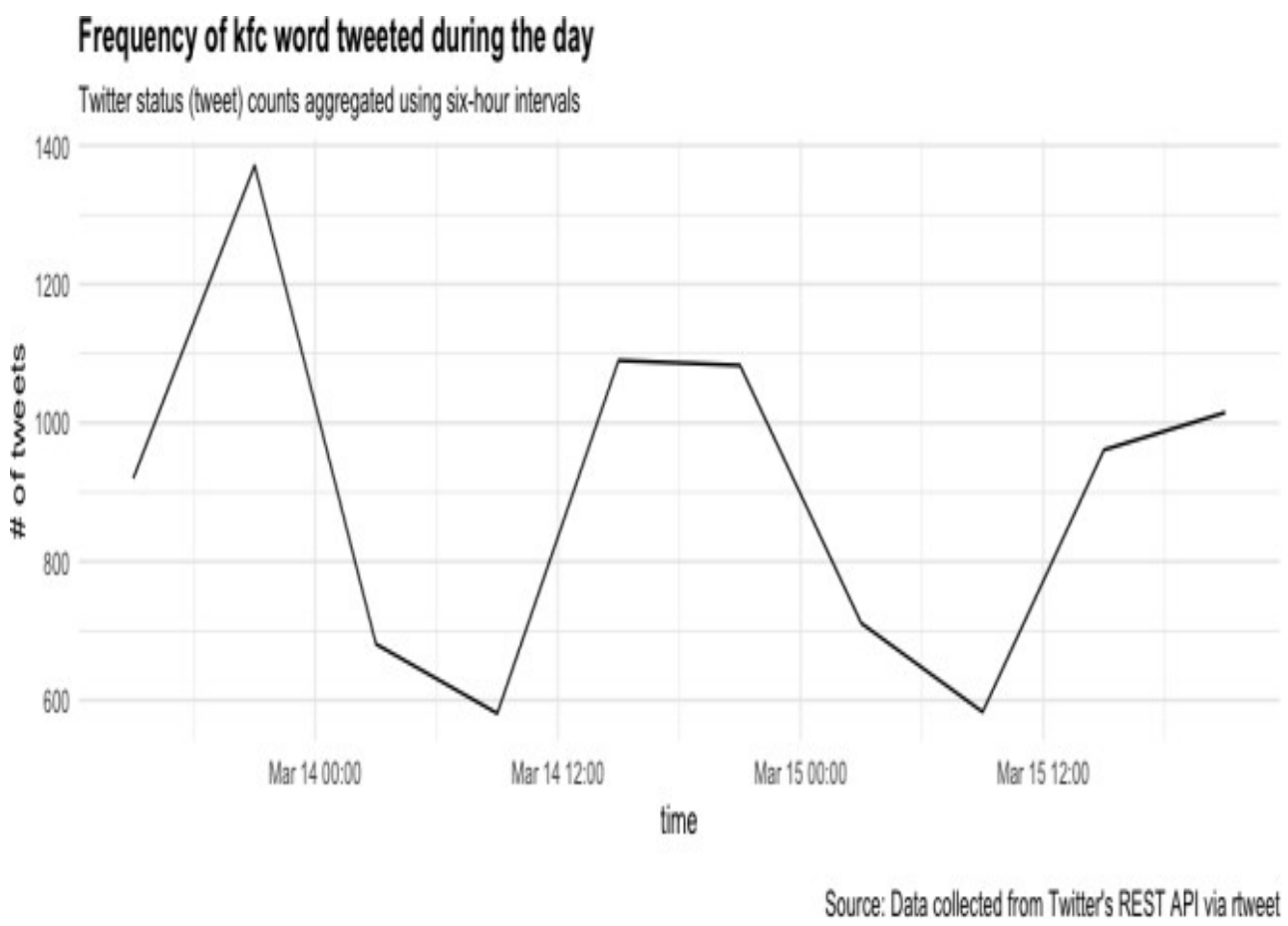

Fig. 2. Frequency of tweets about KFC

In addition, to get more sense about the data we created word clouds using "wordcloud" package in R, generating word cloud from text gives more sense about the most frequently words used in tweets about a specific topic (see Fig. 3). (a)

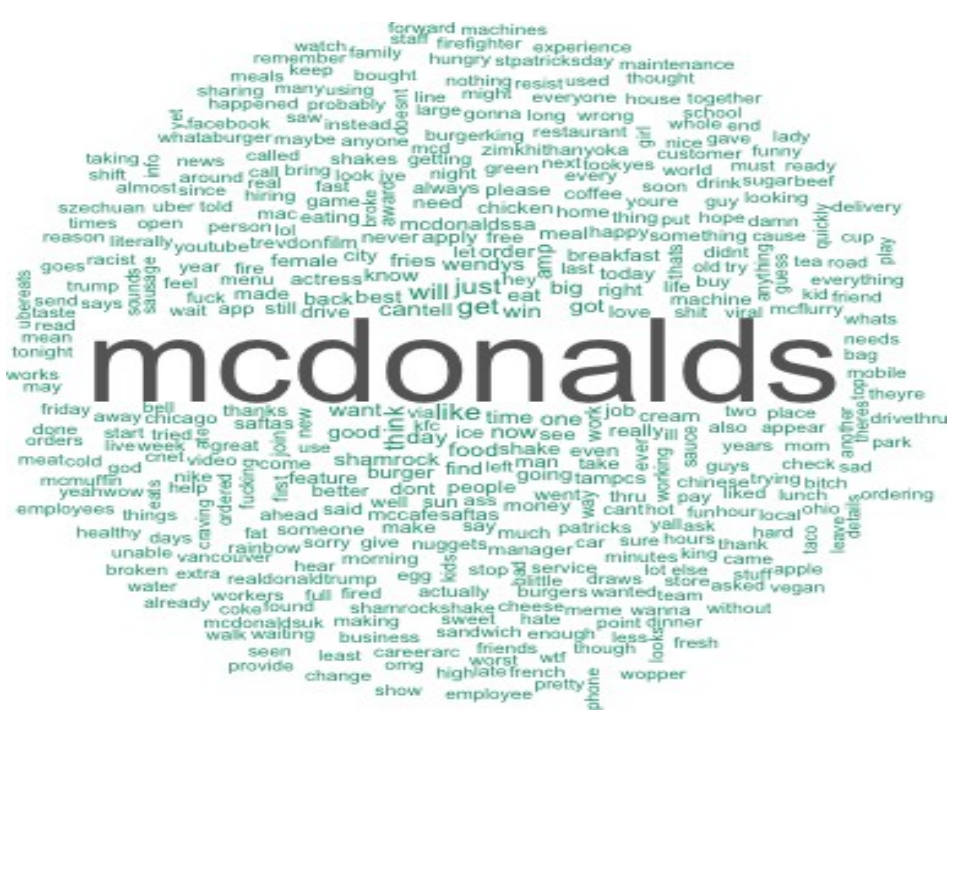

(a)

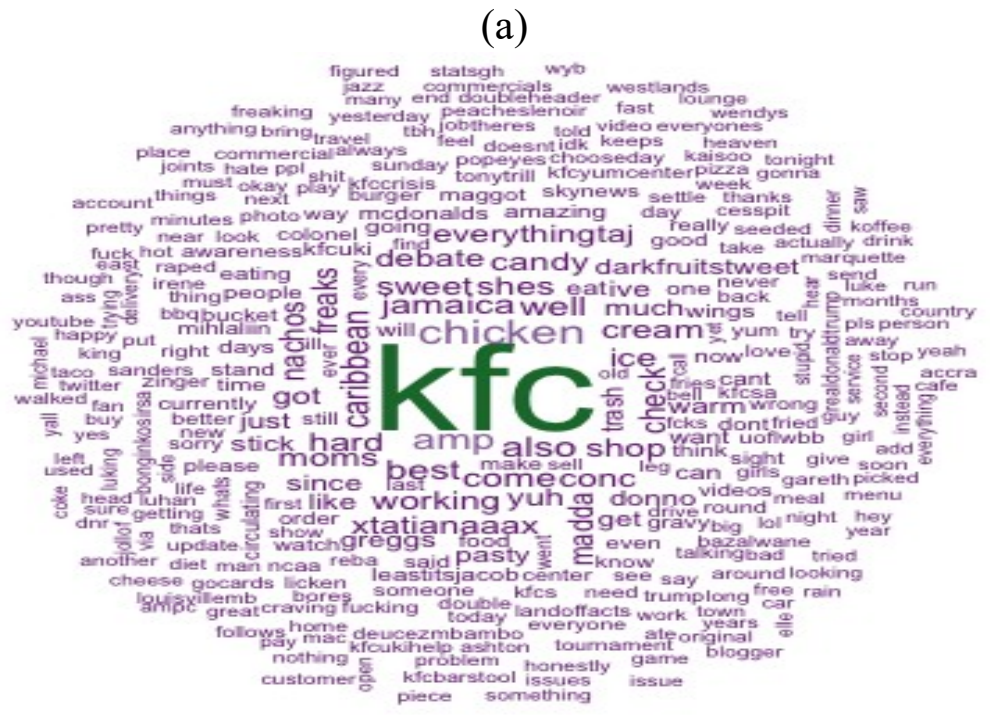

(b)

Fig. 3. Word cloud from text

## A. *Preprocessing*

As the text is highly dimensioned unstructured data, it has to be cleaned and prepared first before analyzing it. Preprocessing the data involves many tasks, depending on type of analysis. In our case we are interested on text only [9-11]: We extracted text from tweets and convert it to data frame, rremoved URLs from text, removed stop words like (the, a, to...), usernames and accounts, removed numbers and unnecessary spaces, removed ppunctuations and cconverting encoding (Emojis) from latin1 to ASCII.

After cleaning the text and removing unnecessary symbols (see Table II), the analysis and mining are performed.

TABLE II. TWEET BEFORE AND AFTER CLEANING

| ***Tweet before Cleaning*** |
|---|
| "@Wendys #1859593 why wendy's is better than McDonalds they messed up all of our orders how u get steak mixed for a egg \xed� " |
| ***Tweet after Cleaning*** |
| "why wendys is better than McDonalds they messed up all orders how u get steak mixed for egg" |

## B. *Model Building*

In this phase, after preparing tweet (removing unnecessary symbols), each tweet was labelled as 1, -1, 0. (That's it: positive, negative, or natural) using unsupervised learning algorithm. Since we do not have pre-classified data, a lexicon-based model used to classify tweets. By using two text files containing a list of positive and negative words, along with more words related to our domain. Each word within each tweet is compared to positive and negative documents in order to find matching words, and classify tweets whether it has more positive or negative words. As a result from this model, the findings are indicated in Table III.

TABLE III. LEXICON BASED CLASSIFICATION

| **Topic** | ***# of tweets*** | ***Positive*** | ***Negative*** | ***Neutral*** |
|---|---|---|---|---|
| McDonald's | 7000 | 2184 | 1589 | 3227 |
| KFC | 7000 | 2076 | 1311 | 3613 |

After that, multiple supervised learning algorithms applied for the purpose of training: Naive Bayes, support vector machine (SVM), maximum entropy, decision tree, random forest and bagging.

- Naïve Bayes: is defined as classifier used to determine the most probable class label for each object.
- Support vector machine: is defined as supervised model, used for classification, regression analysis.
- Maximum entropy: is a classifier used for large variety of text classification.
- Decision tree: are flexible algorithms used to assign label based on the highest score. Random forest: is a supervised algorithm for constructing multiple decision tree.
- Bagging: is a classifier used to taking multiple random samples and use each sample separately to construct a prediction model.

## IV. RESULTS AND DISCUSSIONS

In this paper, data extracted directly from Twitter API were used to train and test the models. A lexicon-based classifier used a manually created lexicon to find the sentiment of each tweet. Our proposed methodology used a novel approach for using both supervised and unsupervised modeling. As a result, the prediction showed improvements in comparison to existing work where a label data is present. Our model combined several algorithms to get the most fit model for our data. Some metrics were used to validate and test the accuracy of each model [12] as follows.

### A. Measurements

- Recall: is defined as number of true positives divided by the number of true positives plus the number of false negatives as indicated in (1).

$$\tau_\pi \mid (\tau_\pi+\phi_\nu) \quad (1)$$

- Precision: is defined as the number of true positives divided by the number of true positives plus the number of false positives as indecated in (2).

$$\tau_\pi \mid (\tau_\pi+\phi_\nu) \quad (2)$$

- Fscore: is a measure of how accurate a model is by using precision and recall following the formula in (3):

*F1_Score = 2 * ((Precision * Recall) / (Precision + Recall))* (3)

### B. Cross validation

In cross validation, the original training data set is divided into four groups, 4-fold cross validation for testing and training.

So we get to know how accurate the model's predictions is when comparing the model's predictions on the validation set and the actual labels of the data points.

After applying validation techniques on the models, the prediction accuracy is found as indicated in Table IV and V .

TABLE IV. ACCURACY RESULT (MCDONALD'S)

| **McDonald's** | | | | |
|---|---|---|---|---|
| **Algorithm** | **Accuracy** | | | |
| | ***Precision*** | ***Recall*** | ***Fscore*** | ***Cross Validate*** |
| Naïve Bayes | 63% | 56% | 51% | 41% |
| SVM | 50% | 33% | 40% | 56% |
| Maxtent | 50% | 22% | 31% | 74% |
| Decision Tree | 80% | 44% | 57% | 54% |
| Random Forest | 33% | 11% | 16% | 58% |
| Bagging | 50% | 33% | 40% | 43% |

TABLE V. ACCURACY RESULT (KFC)

| **KFC** | | | | |
|---|---|---|---|---|
| **Algorithm** | **Accuracy** | | | |
| | ***Precision*** | ***Recall*** | ***Fscore*** | ***Cross Validate*** |
| Naïve Bayes | 41% | 37% | 55% | 45% |
| SVM | 67% | 67% | 67% | 60% |
| Maxtent | 58% | 78% | 67% | 78% |
| Decision Tree | 55% | 67% | 60% | 54% |
| Random Forest | 62% | 89% | 73% | 57% |
| Bagging | 70% | 78% | 74% | 68% |

In addition, testing data of several supervised algorithms showed that Maxent (Maximum entropy) was the best mod(e1l)for both KFC and McDonalds data. As a result of using cross validation as the indicator among other metrics. Also, there was slight difference between the number of positive or negative for both McDonalds and KFC (see Table III). However, more people liking and disliking McDonalds in their tweets. Whereas KFC has more neutral tweets.

## V. CONCLUSION

Sentiment analysis is a field of study for analyzing opinions expressed in text in several social media sites. Our proposed model used several algorithms to enhance the accuracy of classifying tweets as positive, negative and neutral. Our presented methodology combined the use of unsupervised machine learning algorithm where previously labeled data were not exist at first using lexicon-based algorithm. After that data were fed into several supervised model. For testing various metrics used, and it is shown that based on cross validation, maximum entropy has the highest accuracy. As a result, McDonalds is more popular than KFC in terms of both negative and positive reviews. Same methodology can be used in various fields, detecting rumors on Twitter regarding the spread of diseases. For future work, an algorithm that can automatically classify tweets would be an interesting area of research.

## ACKNOWLEDGMENT

We would like to show our gratitude to all participants involved in this work at CCIS, PNU.